\documentclass[conference,compsoc]{IEEEtran}
\ifCLASSOPTIONcompsoc
  \usepackage[nocompress]{cite}
\else
  \usepackage{cite}
\fi
\ifCLASSINFOpdf
   \usepackage[pdftex]{graphicx}
   \graphicspath{{./figures/}}
   \DeclareGraphicsExtensions{.pdf,.jpeg,.png}
\else
   \usepackage[dvips]{graphicx}
   \graphicspath{{./figures/}}
   \DeclareGraphicsExtensions{.eps}
\fi
\usepackage{amsmath}
\usepackage{url}

\usepackage{pythonhighlight}

\begin{document}
%
\title{2CP: Decentralized Protocols to Transparently Evaluate Contributivity in Blockchain Federated Learning Environments}

\author{
\IEEEauthorblockN{Harry Cai}
\IEEEauthorblockA{Imperial College London, UK\\
Vodafone\\
Email: harry.cai@hotmail.com}
\and
\IEEEauthorblockN{Daniel Rueckert}
\IEEEauthorblockA{Imperial College London, UK\\
Technical University of Munich, Germany\\
Email: d.rueckert@imperial.ac.uk}
\and
\IEEEauthorblockN{Jonathan Passerat-Palmbach}
\IEEEauthorblockA{Imperial College London, UK\\
ConsenSys Health\\
Email: j.passerat-palmbach@imperial.ac.uk}
}


%


\maketitle

\begin{abstract}

Federated Learning harnesses data from multiple sources to build a single model. While the initial model might belong solely to the actor bringing it to the network for training, determining the ownership of the trained model resulting from Federated Learning remains an open question. In this paper we explore how Blockchains (in particular Ethereum) can be used to determine the evolving ownership of a model trained with Federated Learning.

Firstly, we use the \textit{step-by-step evaluation} metric to assess the relative contributivities of participants in a Federated Learning process. Next, we introduce \textit{2CP}, a framework comprising two novel protocols for Blockchained Federated Learning, which both reward contributors with shares in the final model based on their relative contributivity. The \textit{Crowdsource Protocol} allows an actor to bring a model forward for training, and use their own data to evaluate the contributions made to it. Potential trainers are guaranteed a fair share of the resulting model, even in a trustless setting. The \textit{Consortium Protocol} gives trainers the same guarantee even when no party owns the initial model and no evaluator is available.

We conduct experiments with the MNIST dataset that reveal sound contributivity scores resulting from both Protocols by rewarding larger datasets with greater shares in the model. Our experiments also showed the necessity to pair 2CP with a robust model aggregation mechanism to discard low quality inputs coming from model poisoning attacks.

\end{abstract}


%
\IEEEpeerreviewmaketitle

\section{Introduction}
Machine Learning allows organisations and researchers to build predictive models from sets of data. In certain circumstances, these predictive models have to be trained on data which is sensitive in nature. For example, models in healthcare settings often have to be trained on patients' data, which is private and confidential. This leads to two problems: firstly, private data is more difficult to acquire because data owners may be reluctant to give away their data or to give consent for their data to be used for machine learning purposes. Secondly, the organisation training the model must keep hold of this data, which introduces security risks and considerations.

\textit{Federated Learning} \cite{mcmahan_communication-efficient_2017} is a technique allowing data owners to contribute their data to the development of machine learning models without revealing it. Broadly speaking, there are two main scenarios in which Federated Learning is useful.

In the \textbf{Crowdsource Setting}, an organisation or team of researchers wish to produce a machine learning model and decide on a set of model hyperparameters and a training protocol. They do not own enough data to train the model themselves, so must draw on the data from multiple outside sources. An example of this scenario would be Google, who train the language model behind Google Keyboard using textual input from users \cite{yang_applied_2018}. 

In the \textbf{Consortium Setting}, multiple organisations and/or individual data owners wish to combine data to train a model that performs better than any model they could train with only their own data. They collectively agree on a set of model hyperparameters and a training protocol. They also agree to share ownership of the resulting model, either by a pre-determined split or based on the value of their individual contributions. \textit{MELLODDY} \cite{noauthor_melloddy_nodate} is an example of this scenario.

\subsection{Data Contributivity for Shared Ownership}
\label{rel:contributivity-considerations}

Parties who collaborate to train a model may agree in advance how to split ownership of it, and any of its resulting utility and revenue. In many of these cases, some parties are able to make greater contributions than others, and would want to see this reflected in their shares. However, the relative value of each party is non-trivial to determine. One must consider many factors together, such as the \textbf{size, quality, representativeness, or novelty} of their local datasets. We can also imagine that a dataset may be poor in isolation, but a valuable addition to other datasets in training a combined model. Conversely, datasets which happen to work well on their own may add little value, or even be a detriment, when combined with other datasets.

In summary, the relative value of collaborating parties' datasets is nearly impossible to determine before training. Furthermore, in Federated Learning settings this problem is compounded by an insurmountable obstacle: parties cannot reveal their data to each other. Therefore, no party has any way to prove the value that their data would bring to the final shared model.

It follows that we should evaluate datasets retrospectively, rather than before the training process. An ideal way of dividing ownership of a trained model (or its profits) would be to split it according to the value that each participant contributes to the final performance of the model, ie. the \textit{contributivity} of their data. This value is objective and can be calculated after the training process is complete. In this case, we do not need to access any participant's data or balance the aforementioned, competing factors. If there were such a protocol, each participant could confidently contribute to the model, with the knowledge that they will be fairly rewarded for it.

\subsection{The Role of Blockchains}

Currently, the most popular Federated Learning protocols \cite{mcmahan_communication-efficient_2017, bonawitz_practical_2017} require a central organiser to execute the training protocol. This is not ideal in the consortium setting - typically, consortia of organisations may not fully trust each other, and may not be able to agree who acts as the central organiser.

Meanwhile, in both crowdsource and consortium settings, participants may need an incentive to participate in the training process. Owners of particularly large or otherwise valuable datasets may also wish to see this reflected in their rewards.

Blockchains allow multiple parties to keep a distributed, immutable and verifiable ledger of records. Meanwhile, blockchain smart contracts can automatically and trustlessly execute pre-determined agreements between them. This work explores how Blockchains - and in particular Ethereum \cite{wood_ethereum_2014} - may be useful to design improved Federated Learning protocols which allow participants to: 1) Pool their resources while fulfilling their own data protection obligations; 2) Allocate shares in the resulting model based on the value of their individual contributions; 3) Enforce these agreements automatically and trustlessly.

\subsection{Summary of Contributions}

\begin{figure}
    \centering
    \includegraphics[width=\columnwidth]{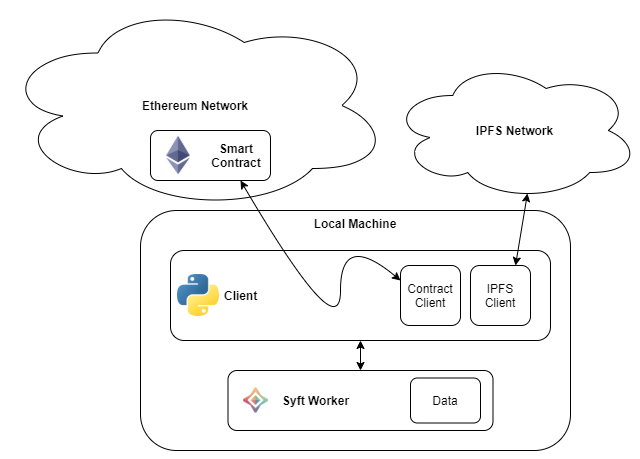}
    \caption{The architecture of the software framework \pyth{2CP}. Double arrows represent communication channels. Local workers are considered to be PySyft \cite{ryffel_generic_2018} workers for the sake of our experiments but could be replaced by other Federated Learning frameworks.}
    \label{fig:client-architecture}
\end{figure}

\begin{itemize}

    \item We explain and advocate \textit{step-by-step evaluation} \cite{noauthor_substrafoundationdistributed-learning-contributivity_2020} as a \textit{contributivity} measure, ie. the relative utility of participants in a Federated Learning process.
    \item We describe two \textbf{new protocols for blockchain-based Federated Learning}: the \textit{Crowdsource Protocol} (Section \ref{ctrb:2cp-crowdsource}) and the \textit{Consortium Protocol} (Section \ref{ctrb:2cp-consortium}). These are each suitable for the crowdsource and consortium settings respectively. Unlike existing protocols \cite{kim_blockchained_2020, nagar_privacy-preserving_2019}, the resulting model's \textbf{ownership is split according to contributivity}.
    \item We develop \pyth{2CP}, a \textbf{software framework} which implements both protocols, and use \pyth{2CP} in Section \ref{sec:experiments} to run experiments using the MNIST dataset to assess the contributivity scores resulting from our protocols. Figure \ref{fig:client-architecture} shows the architecture of \pyth{2CP}.
    

\end{itemize}

\section{Related Work}
\subsubsection*{Data Contributivity} \label{rel:step-by-step}

In an open-source repository, Substra explore potential approaches for measuring the contributivity of participants' data \cite{noauthor_substrafoundationdistributed-learning-contributivity_2020}. Among the approaches they recommend,  \textbf{step-by-step evaluation} is the only one which is possible in a Federated Learning setting, and therefore the one we will base this work upon.

To get the contributivity score for client $A$ using step-by-step evaluation, we define a performance metric $v$ (eg. negative test loss) and record the marginal performance gain $A$ makes to the model $M$ in each iteration $i$, and sum these gains over every iteration in the training process as per Equation \ref{eq:contributivity}. 

\begin{equation}\label{eq:contributivity}
    C(A) = \sum_i \bigg( v(M_i) - v(M_{i+1}^A) \bigg)
\end{equation}

\subsubsection*{Blockchains for Decentralised Federated Learning}

There are proposals for decentralised Federated Learning using Blockchains, with clients submitting model updates 
as transactions to a blockchain \cite{kim_blockchained_2020, nagar_privacy-preserving_2019}. Storing the raw bytes of the models on-chain would be prohibitively expensive, so clients first upload them to IPFS (Inter Planetary File System) \cite{benet_ipfs_2014} then record the CIDs (Content Identifiers - ie. addresses) on-chain instead.

Rather than rely on a central server (which could stall the entire process if it drops out or acts maliciously), clients perform the mean aggregation step at each training round independently and arrive at the same result.

The papers suggesting this architecture use specially designed blockchains, but we replicate them using Ethereum Smart Contracts.

\subsubsection*{Blockchains to Incentivise Federated Learning}

An alternative use of Blockchains in Federated Learning is to use the Blockchain network to record reputation values for each data owner \cite{kang_incentive_2019}. The Federated Learning process employs a model poisoning defence scheme, which detects and rejects malicious updates. These decisions are recorded on the blockchain, and values for \textit{belief}, \textit{disbelief} and \textit{uncertainty} calculated for each client. In future rounds, clients are chosen based on these values, and they earn a set reward for their contribution (if it is not classified as malicious).

Rather than with a flat rate, we intend to reward each update based on its contributivity to model performance.


\section{The Crowdsource Protocol} \label{ctrb:2cp-crowdsource}


For the scenario of the Crowdsource Protocol, we suppose that Alice (the \textit{evaluator}) has a high quality, well distributed and highly representative dataset for a machine learning task. Her dataset is not large enough to train an effective model for this task, but it is sufficient as a test set to evaluate a trained model. Bob, Carol and others (the \textit{trainers}) own suitable datasets to train Alice's model but they are not willing to share them. They are willing to help Alice train a model using Federated Learning, but want to be fairly rewarded for their contributions. They are, of course, unable to reach consensus on how rewards should be split between them. Alice therefore calls upon Bob, Carol et al to train her model using the \textit{Crowdsource Protocol}. 

The protocol goes as follows:

\textbf{1)} Alice deploys the Crowdsource smart contract to the Ethereum blockchain, creates the genesis model, uploads it to IPFS then records its CID on the contract. She also sets the duration, in number of seconds, for each training round. Alternatively, anyone else with an Ethereum wallet could do this, then set Alice's address as the evaluator. Either way, all parties can verify that the contract has been set up correctly and refuse to participate until it has.

\textbf{2)} Once Alice's transactions have been confirmed, the others can see the CID of the genesis model and proceed to download it from IPFS.

\textbf{3)} Using their own data, the trainers each run iterations of training, upload their updated models to IPFS then record their CIDs on the contract. This all needs to be done within the duration of a single training round. The trainers then wait until the next training round starts.

\textbf{4)} On commencement of the next training round, the trainers can see all the CIDs of the updates submitted in the previous training round. They each download these model updates from IPFS and calculate the mean aggregate of them. This is done independently, and they all arrive at the same result.

\textbf{5)} The trainers start from their aggregate (this is the considered the \textit{global model} at the current training round), and repeat the previous steps 3-4 for either a pre-determined number of rounds, or until Alice is satisfied with the global model performance (as evaluated on her own data).

\textbf{6)} The smart contract contains a full history of model updates in each round. Alice can now download all of these and calculate the contributivity of each of them. She assigns a number of \textit{tokens} to each update on the smart contract, as determined by their contributivity. Each token represents a unit of positive contribution to the model. Note that these tokens do not represent financial value nor affect model governance.

The smart contract also contains a record of the model updates each address has made. The tokens owned by each address is the sum of the tokens assigned to each of their updates. Each client's shares in the final model is determined by the proportion of tokens they own.

\begin{figure}
    \centering
    \includegraphics[width=\columnwidth]{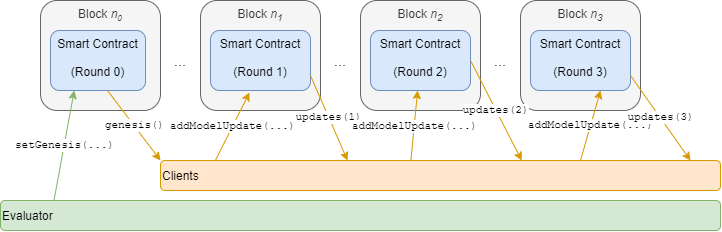}
    \caption{Evaluator and clients interactions with the smart contract during the training process. Each block number $n_i$ denotes the first block with a timestamp that falls within the time range for the training round $i$.}
    \label{fig:fl-blocks-training}
\end{figure}

\begin{figure}
    \centering
    \includegraphics[width=\columnwidth]{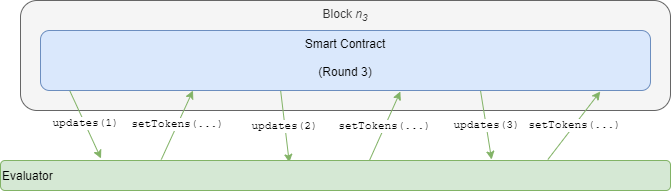}
    \caption{Evaluator interactions with the smart contract during the evaluation process. In this case the evaluator is retrospectively evaluating all updates up to and including the third training round, but they could also evaluate updates as each training round finishes.}
    \label{fig:fl-blocks-evaluate}
\end{figure}

Figure \ref{fig:fl-blocks-training} shows how the evaluator and clients interact with the smart contract. Note that at each block, the smart contract contains the full history of the training process up to that training round. Evaluation can therefore be done in parallel with the training process, retrospectively after the training process, or somewhere in between. The evaluator does not have a time limit to submit their results and does not stall the training process.






We now need an objective performance metric $v$ for the trained model. Our metric is to evaluate the loss of the trained model against a varied, representative and highly accurate holdout test set. The test set must be unavailable to clients and trainers, and used only after the training process is complete, ie. can yield no further changes to the model. In the context of our scenario, Alice's dataset would be used as the holdout test set. Outside of the Crowdsource setting, such an ideal test set is unlikely to exist, and we cannot use the Crowdsource Protocol.

With the performance metric established, it remains to choose a method for determining each participant's contributivity $C$ towards the final model performance $v(M)$. \textbf{Step-by-step evaluation}, as described in Section \ref{rel:step-by-step}, is a suitable method for this. Unlike the other data valuation methods, it can be done in the Federated Learning setting, scores can be calculated retrospectively and do not require the training of any additional models. The computational costs for step-by-step evaluation also increase linearly with the number of clients, in contrast to idealistic approaches based on the Shapley value \cite{shapley_notes_1951}.

We assume that clients' cost of computation is relatively small compared to the value of their data, so every client is incentivised to train on their entire dataset in each iteration, in order to maximise expected performance and reward.

\section{The Consortium Protocol}\label{ctrb:2cp-consortium}

The scenario of the Consortium Protocol sees Alice, Bob, Carol et al each own datasets and form a \textit{consortium}, collaborating to train a Federated Learning model. As a consortium of equals, there is no natural candidate for the \textit{evaluator} role, and hence they cannot use the Crowdsource Protocol from Section \ref{ctrb:2cp-crowdsource}. However, they would still like to split ownership of the resulting model according to the value of their contributions. They therefore make use of the \textit{Consortium Protocol}.


The method for calculating data contributivity described in Section \ref{ctrb:2cp-crowdsource} relies on the existence of a holdout test set by which to evaluate models. In the Consortium setting such a holdout set does not exist. Furthermore, we assume that clients cannot be trusted to set aside any of their data to produce a combined holdout set. This is because we would expect clients to `cheat' and train the model using their holdout data, knowing that this maximises their own contributivity score. As the data used to train local updates is kept secret, they would not be caught if they do this. We also cannot use round-robin based schemes where, for example, clients evaluate one round each in turns - again, because clients would use the same data for both training and evaluation. We therefore propose a new scheme, labelled as \textbf{Parallel Cross Validation}, where clients can use their entire datasets both for training and evaluation.

\begin{figure}
    \centering
    \includegraphics[width=\columnwidth]{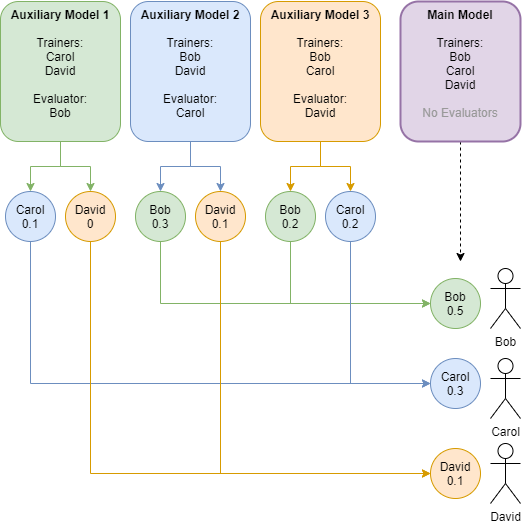}
    \caption{Parallel cross-validation to determine contributivities of a model with no evaluator.}
    \label{fig:parallel-cross-validation}
\end{figure}

Parallel cross validation draws inspiration from the \textit{k-fold cross-validation} technique, where $k=N$ and $N$ is the number of clients. Each round, every client trains $N$ \textit{auxiliary models} (one for each coalition of $N-1$ clients), and evaluates the one model trained by the coalition of all other clients excluding it. Every client also runs a training round on the \textit{main model}, which is trained by all $N$ clients together. The contributivity of each client towards the main model is calculated by summing their contributivities across the auxiliary models. An illustration of the process is shown in Figure \ref{fig:parallel-cross-validation}.

For a consortium of $N$ members to run the Consortium Protocol, they essentially need to run the Crowdsourcing Protocol $N+1$ times, either sequentially or in parallel. In each of the first $N$ of these processes, a different coalition of $N-1$ members trains an auxiliary model, and the remaining member acts as the evaluator. The evaluator uses their data as the holdout set and calculates the contributivity of each trainer. In the final process, all $N$ members train the main model. Remember that we do not have a holdout set by which to evaluate the main model, or to calculate contributivities. Instead, we estimate its performance by averaging the performances of the other $N$ auxiliary models. The contributivity of each trainer is the sum of their contributivities towards the $N-1$ auxiliary models they helped to train.

The Consortium Protocol exposes two limitations compared to the Crowdsource Protocol: \textbf{1)} Unlike the Crowdsource Protocol, the Consortium Protocol is a closed process and participants must be determined beforehand. \textbf{2)} The number of models that need to be trained increases linearly with the number of trainers. All but one of these models are used only for contributivity calculations.





\begin{figure}
    \centering
    \includegraphics[width=\columnwidth]{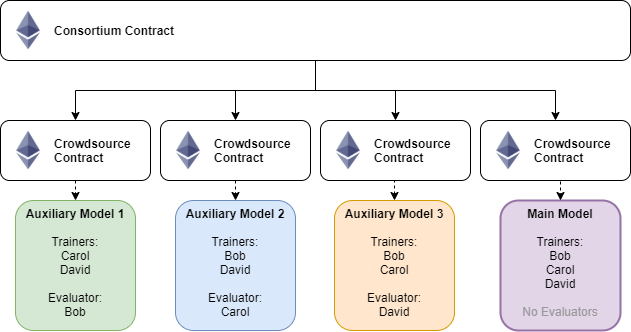}   
    \caption{Suppose Alice, Bob and Carol are running the Consortium Protocol. Their Consortium contract orchestrates 4 Crowdsource contracts, which take charge of a model each.}
    \label{fig:consortium-contracts}
\end{figure}

\section{Experiments}
\label{sec:experiments}

\textit{MNIST} \cite{lecun_gradient-based_1998} is a labelled dataset of handwritten digits. All images are monochrome and $28 \times 28$ pixels. The dataset is split into a \textit{Train Set} of 60,000 images, and a \textit{Test Set} of 10,000 images.

In this section we run a series of experiments using MNIST, each with both protocols, as a validation of our contributivity metric. We present a small selection of results for illustrative purposes in the interest of space. The full set of results will be made available in the \pyth{2CP} repository. For Crowdsource, the entire Test Set is used as the holdout test set (belonging to Alice), while the Train Set is split between trainer clients (Bob \textit{et al}) in various ways. For Consortium, the Train Set is split between trainer clients in the same way, while the Test Set is not used. In all tests, the trainers ran 5 training rounds in which they trained the model using their whole dataset for 1 epoch, with a batch size of 32 and a learning rate of 0.01. In almost all tests, the final global model would achieve an accuracy rate of at least 0.9 on the test set.

In our experiments, almost all tokens are given in the first few rounds. This is likely because our models converge almost fully within two rounds. We also observe very little difference in the relative shares of tokens given to clients between training rounds. A concern expressed about step by step evaluation is that scores might fluctuate wildly between rounds \cite{substra_foundation_value_nodate}, but we have not observed this behaviour.

\begin{figure}
    \centering
    \includegraphics[width=0.49\columnwidth]{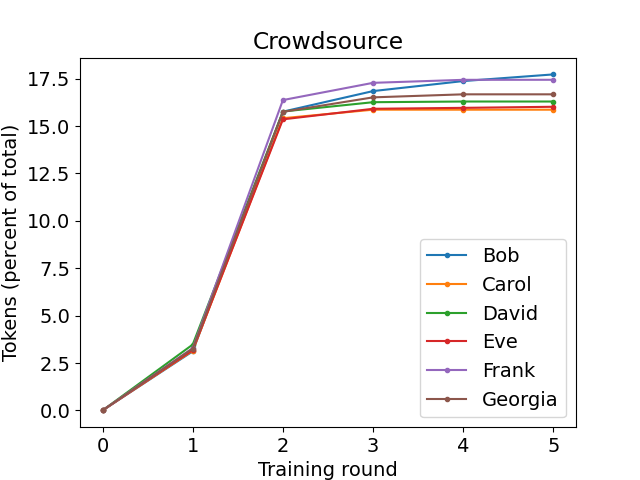}
    \includegraphics[width=0.49\columnwidth]{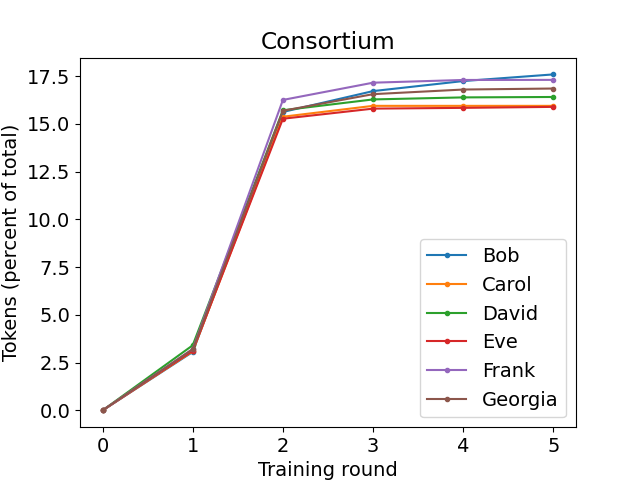}
    \caption{\textit{MNIST Test A.} Results for the Crowdsource Protocol (left) and Consortium Protocol (right) with 6 clients of approximately equal value. As expected, all 6 clients are rewarded with similar token counts. We also see that both protocols yield near identical results, despite using different evaluation metrics.}
    \label{fig:mnist-test-a}
\end{figure}

In \textbf{MNIST Test A}, we split the Train Set randomly and equally between the trainer clients. One therefore expects that each trainer has similar, IID datasets with similar utilities. We vary the number of trainer clients between 2 and 7. As expected, in all cases the token count is split almost equally between clients, reflecting their equal contributions. Remarkably, the Consortium Protocol produces near identical results to Crowdsource at each round, despite its lack of access to Alice's test set. Figure \ref{fig:mnist-test-a} shows one such example.

\begin{figure}
    \centering
    \includegraphics[width=0.49\columnwidth]{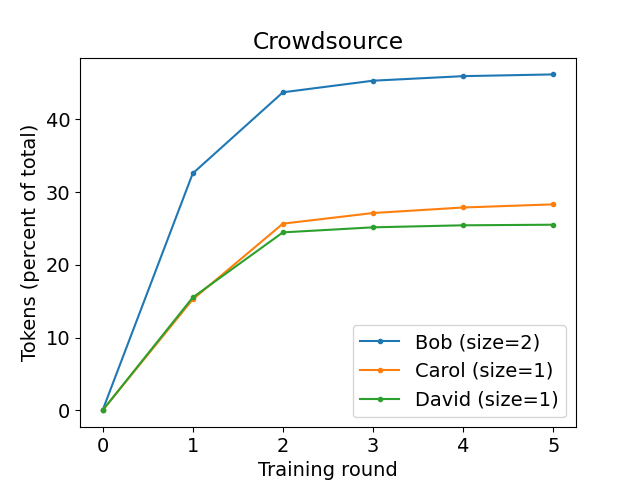}
    \includegraphics[width=0.49\columnwidth]{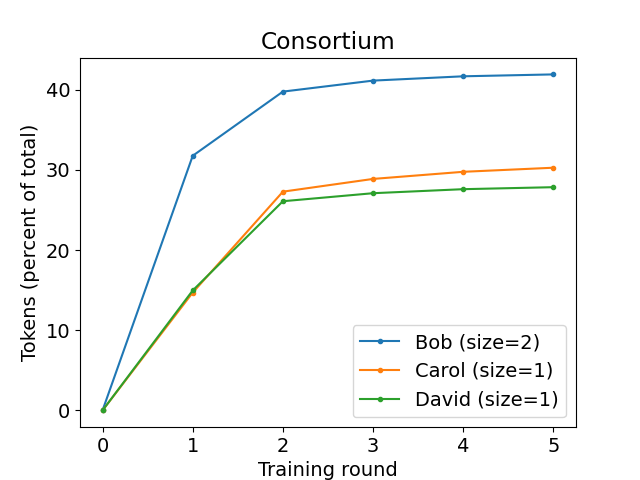}
    \includegraphics[width=0.49\columnwidth]{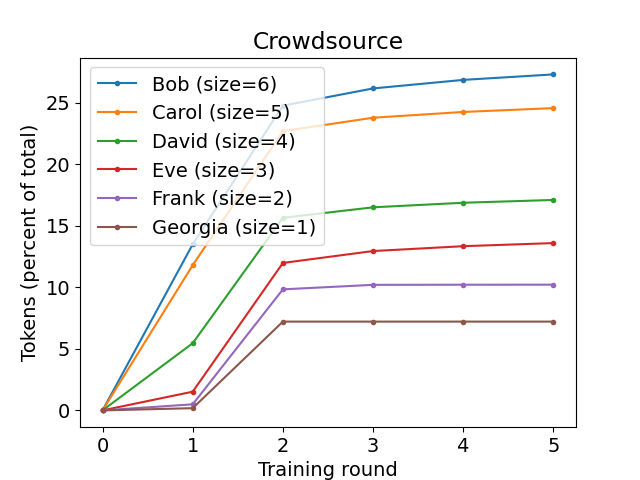}
    \includegraphics[width=0.49\columnwidth]{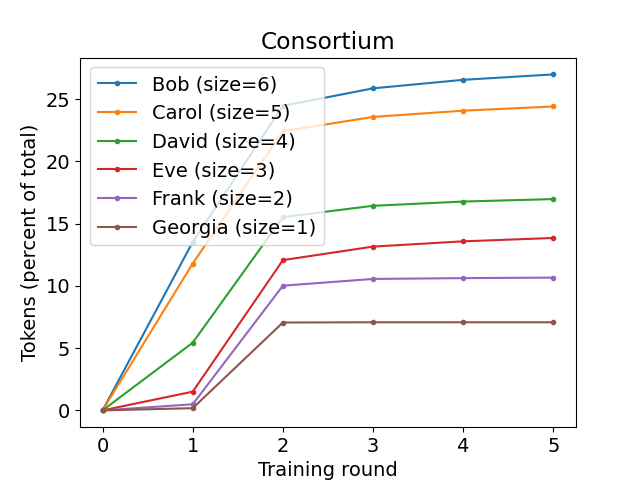}
    \caption{\textit{MNIST Test B.} A selection of results for the Crowdsource Protocol (left column) and Consortium Protocol (right column). In the first scenario (top row), the MNIST Train Set is split randomly between Bob, Carol and David in the ratios 2:1:1. Consequently, Bob is rewarded with a higher token count than Carol and David, who have similar token counts. In the second scenario (bottom row), the Train Set is split randomly between the trainers in the ratio 6:5:4:3:2:1. The resulting token counts reflect the dataset sizes.}
    \label{fig:mnist-test-b}
\end{figure}

In \textbf{MNIST Test B}, the Train Set is split randomly, but in varying ratios. This experiment is intended to isolate the effect of having a smaller or larger dataset on the final token count. As expected, figure \ref{fig:mnist-test-b} shows that larger datasets are better rewarded. In all such experiments, the results from the Crowdsource and Consortium Protocols match each other closely.

\begin{figure}
    \centering
    \includegraphics[width=0.49\columnwidth]{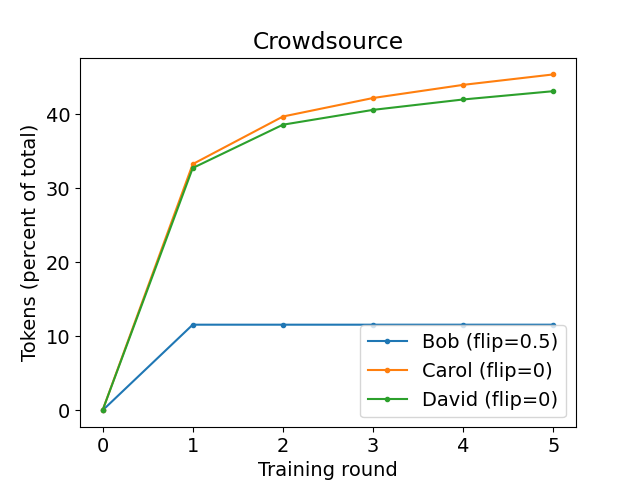}
    \includegraphics[width=0.49\columnwidth]{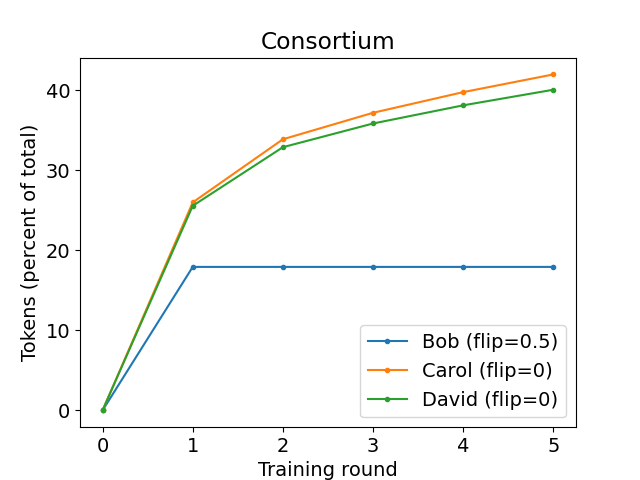}
    \includegraphics[width=0.49\columnwidth]{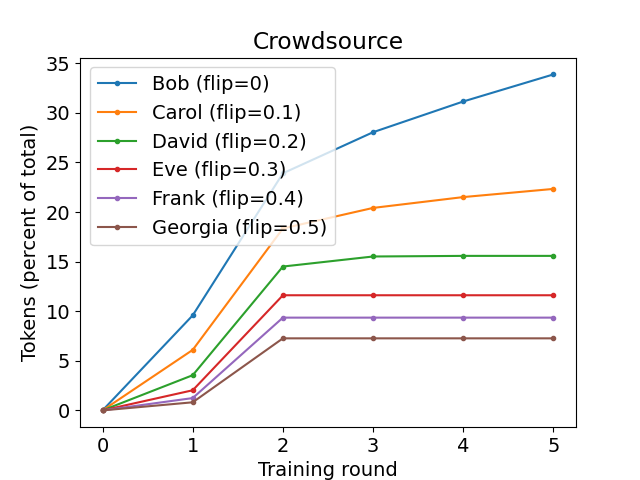}
    \includegraphics[width=0.49\columnwidth]{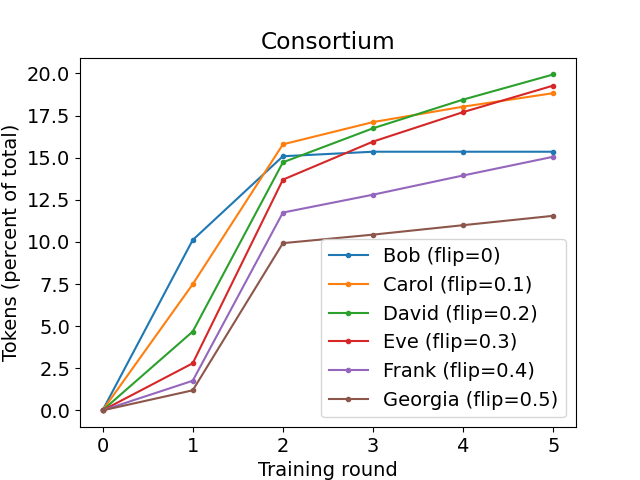}
    \includegraphics[width=0.49\columnwidth]{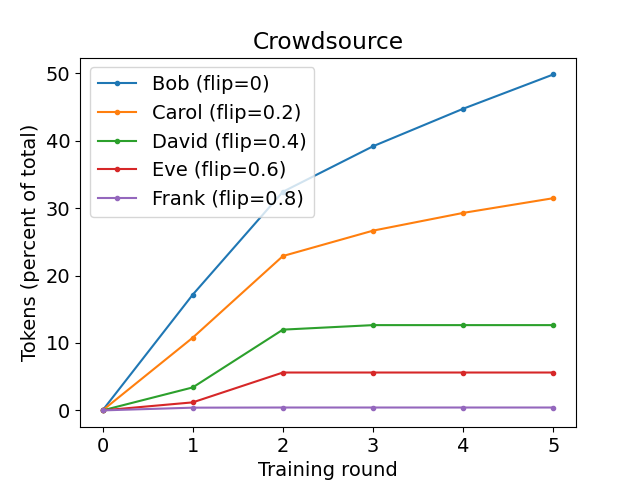}
    \includegraphics[width=0.49\columnwidth]{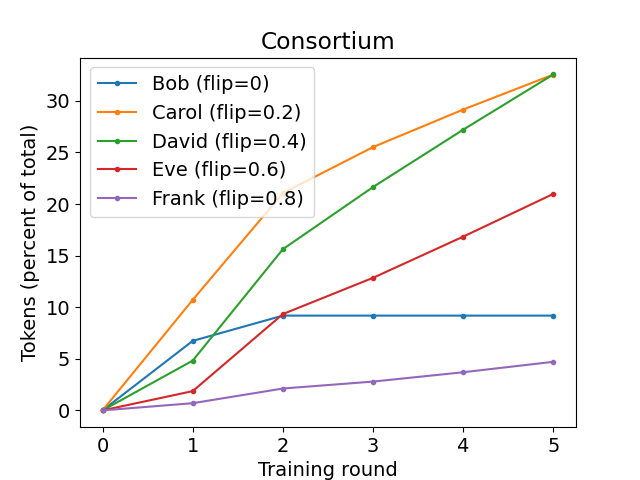}
    \caption{\textit{MNIST Test C.} A selection of results for the Crowdsource Protocol (left column) and Consortium Protocol (right column). The Train Set is split equally and randomly between clients, who each replace a proportion $p$ of their labels with random ones. Crowdsource is able to distinguish between differing $p$ values and penalises them accordingly. Consortium correctly penalises large values of $p$, but it overvalues clients with slightly imperfect datasets ($0<p<0.5$) at the expense of perfect ones. This may be a model poisoning effect from the imperfect datasets.}
    \label{fig:mnist-test-c}
\end{figure}

Finally, \textbf{MNIST Test C} splits the Train Set as in A, but the trainers each replace a proportion $p$ of their labels with a random one, simulating a label flipping attack to try poisoning the model. One therefore hopes that clients with higher $p$ values receive fewer tokens. In Figure \ref{fig:mnist-test-c}, we can see that for Crowdsource, more label flipping does lead to fewer tokens, as it holds up less favourably to the unmodified holdout set. However, Figure \ref{fig:mnist-test-c} also shows that for Consortium, a flip probability $0<p<0.5$ actually increases eventual token shares at the expense of unmodified trainers. This quickly drops off to zero for $p>0.5$. Among our experiments, this effect is greatest when there are many flipped labels overall, and $0.2<p<0.3$ is rewarded most. In these scenarios, these clients essentially amount to a Sybil attack for untargeted model poisoning \cite{kairouz_advances_2019}. The interpretation is that they have succeeded in diverting the model parameters to a sub-optimal local minimum, where unmodified trainers are punished for attempting to bring the model parameters away. Note that the early training rounds match the token shares from Crowdsource, which supports this theory. The Consortium measure should however be adjusted to avoid these situations.

This final experiment highlights the limitations of Federated Averaging. The Consortium protocol would thus benefit from being used in conjunction with a robust aggregation mechanism dismissing malicious updates as in \cite{munoz-gonzalez_byzantine-robust_2019}.

\section{Conclusion}
In this paper, we have introduced and implemented two protocols to calculate the contributivity of each participant's data set in a Federated Learning network. Both protocols are decentralised and maximise the fairness in the calculation.

Our experiments showed that the contributivity scores resulting from the Crowdsource Protocol was sound when computed for the MNIST dataset. Clients with larger or higher quality datasets were rewarded with higher shares in the final model. Clients with less to contribute, but whose contributions would still be positive, are still given rewards for their participation. This result was obtained with a high quality holdout test set. The Consortium Protocol correctly rewarded larger datasets with higher scores, but our experiments revealed situations where low quality datasets were given higher scores than perfect ones.


At the time of writing, our protocols and \pyth{2CP} are still under active development. It will be interesting to evaluate the two protocols when trainers have deliberately different and unique distributions of data, instead of being randomly sampled. Similarly, we want to understand the impact of privacy-preserving mechanisms such as Differential Privacy on the quality of the model updates, and hence on the contributivity scores and shares in the final model.

    
    
    

\pyth{2CP} does not yet support verified evaluation and verified training as described in this paper. The contributivity scheme only produces fair results if all clients follow the protocol honestly, or semi-honestly (honest but curious). Malicious clients can freely push bad and/or fake model updates, making the training process highly vulnerable to both untargeted and targeted model poisoning attacks \cite{kairouz_advances_2019}. Future work will consequently focus on designing a verifiable evaluation and training procedure, as well as designing a penalty scheme to use in conjunction. From a technical standpoint, we aim to optimise the \pyth{2CP} framework for gas costs and scalability to make it suitable for public deployment on Ethereum and IPFS.

\end{document}